\documentclass{llncs}

\usepackage{graphicx}
\usepackage{amsmath}
\usepackage{amssymb}
\usepackage[ruled,linesnumbered,vlined]{algorithm2e}
\usepackage{xcolor}
\usepackage{tabularx}
\usepackage{tabu}
\usepackage{multirow}
\usepackage{colortbl}
\usepackage{url}
\usepackage{hyperref}
\usepackage{caption} 
\usepackage{mathtools}

\usepackage{fancyvrb}
\DefineVerbatimEnvironment{shell}{Verbatim}{commandchars=\%\{\},formatcom=\setcounter{prompt}{0}}
\newcounter{prompt}

\title{Collaborative Management of Benchmark Instances and their Attributes}
\author{Markus Iser, Luca Springer, Carsten Sinz}
\institute{
  Karlsruhe Institute of Technology (KIT), Germany\\
  \url{{markus.iser,carsten.sinz}@kit.edu}\\
  \url{luca.springer@student.kit.edu}
}

\begin{document}

\maketitle

\begin{abstract}
Experimental evaluation is an integral part in the design process of algorithms.
Especially for hard combinatorial problems, as they occur in the areas of constraint programming and SAT solving, publicly available benchmark instances are widely used to evaluate new methods.
In this paper, we present our system GBD Tools which simplifies experimental evaluation with a searchable instance catalog, databases of instance features, extraction of instances with specific attributes, and analysis of results with augmented runtime data.
Central to our tool is the specification of a benchmark instance identifier.
In this paper, we exemplarily define such an identifier, GBD Hash, for DIMACS CNF.
GBD Hash simplifies the integration of different SAT benchmark instance repositories and facilitates the consolidation of solver runtimes, benchmark instances and instance meta-data.
We expect GBD Tools to be helpful for researchers who empirically evaluate methods for hard combinatorial problems. 
\end{abstract}

\section{Introduction}

The experimental evaluation of implementations plays a crucial role in algorithmic discovery. 
Implementations of algorithms that solve hard combinatorial problems, e.g., SAT solvers, are usually evaluated with publicly available benchmark instances. 
The analysis of such runtime data can support and inspire hypotheses and theories. 
For the analysis of runtime data, it is important to take additional information about the benchmark instances into account. 
But meta features, such as instance origin, solution, model count, etc. are not easily available. 

Experiments with SAT solvers usually fall back on sets of instances which have been compiled for the annual SAT competitions and most information about the benchmark instances can be found in the respective competition proceedings~\cite{SC2020}. 
\textsf{SatLib} is an early structured public collection of benchmark instances for experiments in SAT~\cite{Hoos:2000:SatLib}. 
\textsf{SatEx} is the first web-based framework for reproducible experimentation and experiment evaluation~\cite{Simon:2001:SatEx}. 

Instance features can be used for instance-specific algorithm selection~\cite{Hoos:2011:SATzilla}. 
\textsf{Aslib} is a library for reproducible training of such runtime prediction models~\cite{Hoos:2016:Aslib}. 
Studies of correlations between solver runtimes and instance features also shed light on the reasons for the feasibility of industrial instances~\cite{Ganesh:2021:HCS}. 
Instance features are also used in empirical studies of proof complexity~\cite{Elffers:2018:Heuristics}, and to reduce redundancy in experimentation~\cite{Manthey:2016:Structure}.

In our workshop paper for Pragmatics of SAT (POS) 2018~\cite{Iser:2018:GBD}, we suggested to bring together several lines of research via instance identification and there we discussed the aspects of a good benchmark instance identifier, and presented use-cases with a proof-of-concept implementation. 
Since then, we finalized a specification of our benchmark instance identifier 
and created \textsf{GBD Tools}. 
Our tool has a wide range of use-cases in experimental research in SAT solving which can be categorized as follows.%

\begin{itemize}
\item GBD is a searchable catalog for instance download 
\item GBD has several databases with readily bootstrapped instance features
\item GBD facilitates experimentation with instances with specific attributes
\item GBD supports the analysis of augmented runtime data
\end{itemize}

In Section~\ref{sec:id} of this paper, we present the definite specification of \textsf{GBD Hash} and elaborate on the reasoning behind it. 
In Section~\ref{sec:system}, we present a system description of \textsf{GBD Tools}. 
In Section~\ref{sec:tool}, we walk through several use-cases of \textsf{GBD Tools}, e.g., the evaluation and visualization of \emph{augmented} runtime data. 
In Section~\ref{sec:web}, we present the web-frontend of \textsf{GBD Tools} for distribution of benchmark instances and data. 
We conclude with Section~\ref{sec:conclusion}.

\section{Benchmark Instance Identification}
\label{sec:id}

\begin{table}[t]
\centering
\small
\extrarowsep=.03em
\setlength{\tabcolsep}{1.5em}
\begin{tabular}{llrr}
\bf Competition & \bf Track & \bf Nominal & \bf Actual \\
\hline
SAT Competition 2011 & MUS 			& 300 & 299 \\
\arrayrulecolor{gray}
\hline
\multirow{2}{*}{SAT Challenge 2012} & Application  & 600 & 596 \\
								 	& Portfolio	& 600 & 599 \\
\hline
SAT Competition 2014 & Application  & 300 & 299 \\
\hline
\multirow{2}{*}{SAT Race 2015}	    & Main			& 300 & 291 \\
							 		& Parallel		& 100 & 96 \\
\hline
\multirow{2}{*}{SAT Competition 2016} & Agile 		& 5000 & 1580 \\
									& Application  & 300 & 299 \\
\hline
\multirow{2}{*}{SAT Competition 2017} & Agile 		& 5000 & 2379 \\
									& Random		& 300 & 294 \\
\hline
SAT Race 2019 		 & Main 		& 400 & 399
\end{tabular}~\\[.9em]
\caption{Diverging \emph{nominal} and \emph{actual} numbers of instances in SAT competition benchmarks}
\label{tab:competitions}
\end{table}

\begin{function}[b]
\DontPrintSemicolon
\KwIn{Benchmark Instance (DIMACS)}
\KwOut{GBD Hash}
Remove Comments and Header\;
Replace Newline and Carriage Returns by Blank Spaces\;
Replace any Sequence of Whitespace by a single Blank Space\;
Fix ``Missing Trailing Zero'' for Last Clause\;
\Return \textsf{md5sum} of the remaining content\;
\caption{GBD Hash(Benchmark Instance)}
\label{func:hash}
\end{function}

When maintaining and distributing benchmark instances, solver runtimes and other instance attributes, the problem of instance identification arises. 
The distributed management of instances and their properties places special demands on instance identification. 

The instances themselves are usually very \emph{large}, and we need a short identifier to manage and distribute instance data independently of the instances. 
Instance filenames sometimes encode meta features, as such they are interesting raw features. 
But instances filenames are easy to change and thus \emph{duplicates} of the same instances with different filenames can be found in many benchmark sets, even in singular SAT Competitions (cf. Table~\ref{tab:competitions}). 
Duplicates can also occur when instance generators produce the same instances in corner-cases of encodings. 
Hashsums of files containing benchmark instances can be used to verify the integrity of files. 
But as an instance identifier they are not suitable, since hashsums are \emph{volatile} regarding small changes to whitepaces, comments, header, or compression formats. 

Function~\ref{func:hash} specifies a benchmark \emph{instance identifier} for \textsf{DIMACS CNF}~\cite{dimacs:1993:cnf}, which creates the hashsum of the \emph{normalized} benchmark instance. 
Thanks to normalization, \textsf{GBD Hash} is invariant to removal and addition of comments as well as (possibly erroneous) header information (Line~1). 
Line-break as well as carriage-return characters are replaced by blank spaces to achieve invariance over several newline encodings of the respective operating systems (Line~2). 
Further whitespace normalization ensures that all sequences of non-space characters are separated by a single blank (Line~3). 
Since the DIMACS standard allows for omitting the last clause's sentinel element~\cite{dimacs:1993:cnf}, we make sure to include a trailing zero also for the last clause (Line~4). 
The such normalized content is used as the input to a hash function for final calculation of the instance identifier (Line~5). 

A possible alternative could be an identifier for the equivalence class of shuffled instances. 
We decided against establishing any kind of variable or clause order invariance, since solver runtimes (which are associated with specific instances) are sensitive to such reordering~\cite{Biere:2018:Scrambling}. 
In Section~\ref{sec:tool}, we show how identifiers of weaker equivalence classes can be included via feature bootstrapping.

\section{GBD Tools -- System Description}
\label{sec:system}

\begin{figure}[t]%
\scriptsize%
\newcommand{\pr}[1]{\ensuremath{\langle #1 \rangle}}%
\newcommand{\tspp}[1]{\ensuremath{\text{'\textsf{#1}'}}}%
\newcommand{\tbs}{\textbackslash}%
\def\-{\raisebox{.75pt}{-}}%
\def\u{\raisebox{.75pt}{\_}}%
\begin{align*}
    \pr{start} ~=~ &\pr{query} \mid \epsilon \\
    \pr{query} ~=~ &\tspp{(}, \pr{query}, \tspp{)} \mid \pr{query}, (\tspp{~and~} \mid \tspp{~or~}), \pr{query} \mid \pr{constraint} \\
    \pr{constraint} ~=~ &\pr{name}, (\tspp{=} \mid \tspp{!=}), \pr{value} \mid 
    	\pr{name}, \tspp{~like~}, [\tspp{\%}], \pr{value}, [\tspp{\%}] \mid \\        
    					 &\tspp{(}, \pr{term}, (\tspp{=} \mid \tspp{!=} \mid \tspp{<} \mid \tspp{>}), \pr{term}, \tspp{)}\\
    \pr{term} ~=~ &\pr{name} \mid \pr{number} \mid 
    	~\tspp{(}, \pr{term}, (\tspp{+} \mid \tspp{-} \mid \tspp{*} \mid \tspp{/}), \pr{term}, \tspp{)} \\
    \pr{name} ~=~ & \pr{letter}, \{ \pr{letter} \mid \pr{digit} \mid \tspp{\u} \}\\
    \pr{number} ~=~ & [\tspp{-}] \pr{digit} \{ \pr{digit} \} [\tspp{.} \pr{digit} \{ \pr{digit} \}] \\
    \pr{value} ~=~ & \{ \pr{letter} \mid \pr{digit} \mid \tspp{\u} \mid \tspp{.} \mid \tspp{-} \mid \tspp{/} \} \\
    \pr{letter} ~=~ &\tspp{a} \mid \tspp{b} \mid \dots \mid \tspp{z} \mid \tspp{A} \mid \tspp{B} \mid \dots \mid \tspp{Z} \\
    \pr{digit} ~=~ &\tspp{0} \mid \tspp{1} \mid \dots \mid \tspp{9}
\end{align*}%
\vspace{-2em}
\caption[GBD Queries]{EBNF Description of GBD Queries}
\label{fig:gbdql}
\end{figure}

\textsf{GBD Tools} is available in the Python Package Index (PyPI).% 
\footnote{\textsf{pip3~install~gbd-tools}}
For contributions, we maintain a public repository on GitHub.% 
\footnote{\url{https://github.com/Udopia/gbd}}
\textsf{GBD Tools} comes with the command-line tool \textsf{gbd}, which has a complex sub-command structure (cf. Table~\ref{tab:gbd-cli}), e.g., for augmented runtime analysis, feature-based instance filtering, or feature bootstrapping (cf. Section~\ref{sec:tool}). 
\textsf{GBD Tools} also comes with \textsf{gbd-server} which runs a web-interface that gives access to instances and data (cf. Section~\ref{sec:web}). 

We maintain databases of instance attributes, e.g., of meta-attributes such as instance family and author, or the feature records used in SATzilla~\cite{Hoos:2011:SATzilla}, or of solver runtimes in SAT competitions. 
GBD databases associate all attributes (including runtimes) with \textsf{GBD Hash} in SQLite or CSV files. 
In order to register a database, its path can be appended in the \texttt{GBD\_DB} environment variable. 

Filtering for specific instances, e.g., of a specific competition or a specific instance family, is facilitated by GBD Queries. 
GBD Queries can also be used in augmented runtime analysis in order to visually highlight instances with specific attributes. 
Figure~\ref{fig:gbdql} depicts the query language of \textsf{GBD Tools} in Extended Backus-Naur Form (EBNF)~\cite{ISO14977}. 
GBD Queries exist to simplify data access for several commands in \textsf{GBD Tools} and internally, they are automatically translated to SQL Queries. 

Table~\ref{tab:gbd-cli} displays \textsf{gbd} sub-commands of first order. 
The command \textsf{get} is one of the most commonly used commands and responsible for information retrieval using a combination of \textsf{GBD Query} and further parameters. 
The commands \textsf{eval} and \textsf{plot} facilitate the inclusion of multiple data-sources for numerical and visual analysis of \emph{augmented} runtime data. 

The \textsf{init} command for feature bootstrapping has an \emph{extensible} sub-command structure with \emph{parallelization} support. 
The sub-command \textsf{gbd init local} registers paths to local benchmark instances.
Notably, \textsf{gbd init local} runs a lot faster if our dedicated Python accelerator module is installed.\footnote{\url{https://github.com/Udopia/gbdhash}}
An intersting new bootstrapping command \textsf{gbd init degree\_sequence\_hash} calculates the hashsum of the sorted degree-sequence of an instance's graph. 
This feature over-approximates the equivalence class of shuffled instances.

Further commands such as \textsf{set}, \textsf{import} or \textsf{create} are used to store data, e.g., solver runtimes, in the database. 
For all commands, information about its parameters can be retrieved by using the respective help, e.g., \texttt{gbd set ---help}.

\begin{table}[t]
\centering
\small
\arrayrulecolor{gray}
\extrarowsep=.1em
\setlength{\tabcolsep}{.7em}
\begin{tabularx}{\linewidth}{Xl}
\multicolumn{2}{l}{\bf Initialization and Bootstrapping}\\
\hline
\textsf{gbd init}$~^\mathsf{improved}$   ~&~ Initialize and/or Bootstrap Database\\
&\\[-.5em]
\multicolumn{2}{l}{\bf Information Retrieval}\\
\hline
\textsf{gbd get}$~^\mathsf{improved}$    ~&~ Query for Instances and/or Features\\
\textsf{gbd info}$~^\mathsf{improved}$   ~&~ Display available Databases and Attributes\\
\textsf{gbd eval}$~^\mathsf{new}$   ~&~ Evaluate Runtime Experiments\\
\textsf{gbd plot}$~^\mathsf{new}$   ~&~ Visualize Augmented Runtime Data\\
&\\[-.5em]
\multicolumn{2}{l}{\bf Information Management}\\
\hline
\textsf{gbd set}$~^\mathsf{improved}$    ~&~ Set Features\\
\textsf{gbd import}$~^\mathsf{new}$ ~&~ Import Features from CSV\\
\textsf{gbd create/delete/rename}$~^\mathsf{improved}$ ~&~ Create/Delete/Rename Attributes\\
\textsf{gbd info\_set/info\_clear}$~^\mathsf{new}$ ~&~ Manage Meta-Attributes
\end{tabularx}~\\[1em]
\caption{Commands of GBD Command Line Interface}
\label{tab:gbd-cli}
\end{table}

% \newpage
\section{Use-Cases: From Experiment to Evaluation}
\label{sec:tool}

In the folling use-cases, we assume \textsf{gbd-tools} to be installed and our meta database\footnote{\url{http://gbd.iti.kit.edu/getdatabase/meta.db}} to be available and registered in the \textsf{GBD\_DB} environment variable. 
In some examples, we assume that instances of the Main tracks of SAT Competitions~2019 and~2020 are available in the directory \texttt{/home/jane/cnf}. 

\subsubsection*{Testing the Setup.}
In order to get information about the configured database, 
the available attributes, or their value-ranges, 
we can use the info command, like in the following example.

\begin{shell}
%prompt gbd info
Database: /home/jane/meta.db
Features: local variables clauses family author ...
\end{shell}

\subsubsection*{Accessing Data.}
Data access with \textsf{gbd get} is the original application of \textsf{GBD Query}. 
\textsf{GBD Query} filters for instances that match the specified constraints. 
In the following example, the command prints hashes of instances of SAT Race~2019 in the first column. 
In the second and third column, it prints their instance family and filename. 

\begin{shell}
%prompt gbd get "competition_track = main_2019" -r family filename
0447371bb8a97e8fe5d3cee6de1db766 diagnosis UTI-20-10p0-sc2009.cnf
...
\end{shell}
The next example shows that \textsf{gbd get} also uses hash values read from stdin. 
This is useful to augment data in hindsight. 

\begin{shell}
%prompt echo fd972d87a0e61efaba7890da38a12f8c | gbd get -r author
fd972d87a0e61efaba7890da38a12f8c frioux
\end{shell}
We can also change the primary key to the degree-sequence hash. 
Like this we can find instances of the same degree-sequence, as can be seen in the following example. 

\begin{shell}
%prompt gbd get -g degree_sequence_hash -r filename
aeee0fc6376c8e15622bec6e9f1e23d0 sted3_0x1e3-147.cnf
af6c1679f40279044b49fc1b18ff6c07 TT7F-33-24B.cnf,TT7F-33-24C.cnf
...
\end{shell}

\subsubsection*{Feature Bootstrapping.}

In order to get experimentation elevated with \textsf{gbd}, we need to register the paths to our locally available instances. 
This \emph{initial} bootstrapping process associates local paths with gbd hashes which need to be calculated. 
This process completes in a reasonable time if the \textsf{gbdhash} accelerator module is installed. 
We can bootstrap local paths in parallel as follows. 

\begin{shell}
%prompt gbd --jobs=8 init local /home/jane/cnf
\end{shell}
The \textsf{init} command is extensible. 
Further sub-commands are available or under development, e.g. dedicated feature extractors.  
Another interesting feature to bootstrap with \textsf{init}, is the \textsf{degree\_sequence\_hash} as a weaker equivalence class identifier. 
Bootstrapping of \textsf{local} paths takes on a special role, since \textsf{local} is a special type of feature that stores paths to the locally available instances.

\subsubsection*{Experimenting.}
Having initialized local paths, we can resolve queries against the special attribute \textsf{local}, which returns instance paths. 
In the following examples, we query for the local paths of instances used in SAT Competition 2020 (1), and a set of available cryptographic instances (2). 

\begin{shell}
%prompt gbd get "competition_track = main_2020" -r local
fc538412229f86bc7016b8bc0aca3924 /home/jane/cnf/g2-T83.2.1.cnf
...
%prompt gbd get "family = cryptography" -r local
e0ebbebd8c2372191ce9425292f241c5 /home/jane/cnf/sha1r17m72a.cnf
...
\end{shell}
The following example shows an additional aspect of \textsf{GBD Query}. 
We use arithmetic expressions to return all instances of the family \emph{uniform-random} of the given minimum \emph{clause/variable ratio}. 

\begin{shell}
%prompt gbd get "family = uniform-random and (clauses / variables) > 4.2"
01c48559d4870a5f3216291060fbeb39
...
\end{shell}

\subsubsection*{Evaluation.}
Runtime data can be imported by creating custom attributes using the \textsf{create}, \textsf{import} and \textsf{set} commands. 
In the following examples, we use the runtime data of the Main track of SAT Competition~2020, which is available on our website.\footnote{\url{http://gbd.iti.kit.edu/getdatabase/sc2020.db}}
With the command \textsf{eval par2}, we can display the PAR-2 scores and the number of solved instances for the specified solvers as follows. 

\begin{shell}
%prompt gbd eval par2 "competition_track = main_2020" -t 5000 
	-r kissat_sat relaxed_newtech
	
kissat_sat:      3926.2 (264 / 400)
relaxed_newtech: 4140.3 (255 / 400)
vbs:             3323.5 (285 / 400)
\end{shell}
The command \textsf{eval comb} calculates the PAR-2 score of the VBS of all tuples of size $k$ drawn from a given set of solvers. 
This function can be used to analyze the orthogonality of solvers with regard to portfolio construction as follows. 

\begin{shell}
%prompt gbd eval comb "competition_track = main_2020" -k 3 -t 5000 
	-r cadical_sc2020 kissat exmaple_padc_dl ... | sort
	
2986.4 kissat_unsat, relaxed_newtech, cadical_sc2020
2987.7 kissat_unsat, relaxed_newtech, cryptominisat_walksat
2988.2 kissat_unsat, relaxed_newtech, kissat_sat
...
\end{shell}

\subsubsection*{Plotting.}
\begin{figure}[t]
\centering
\includegraphics[width=.87\linewidth, trim=150 10 150 10, clip]{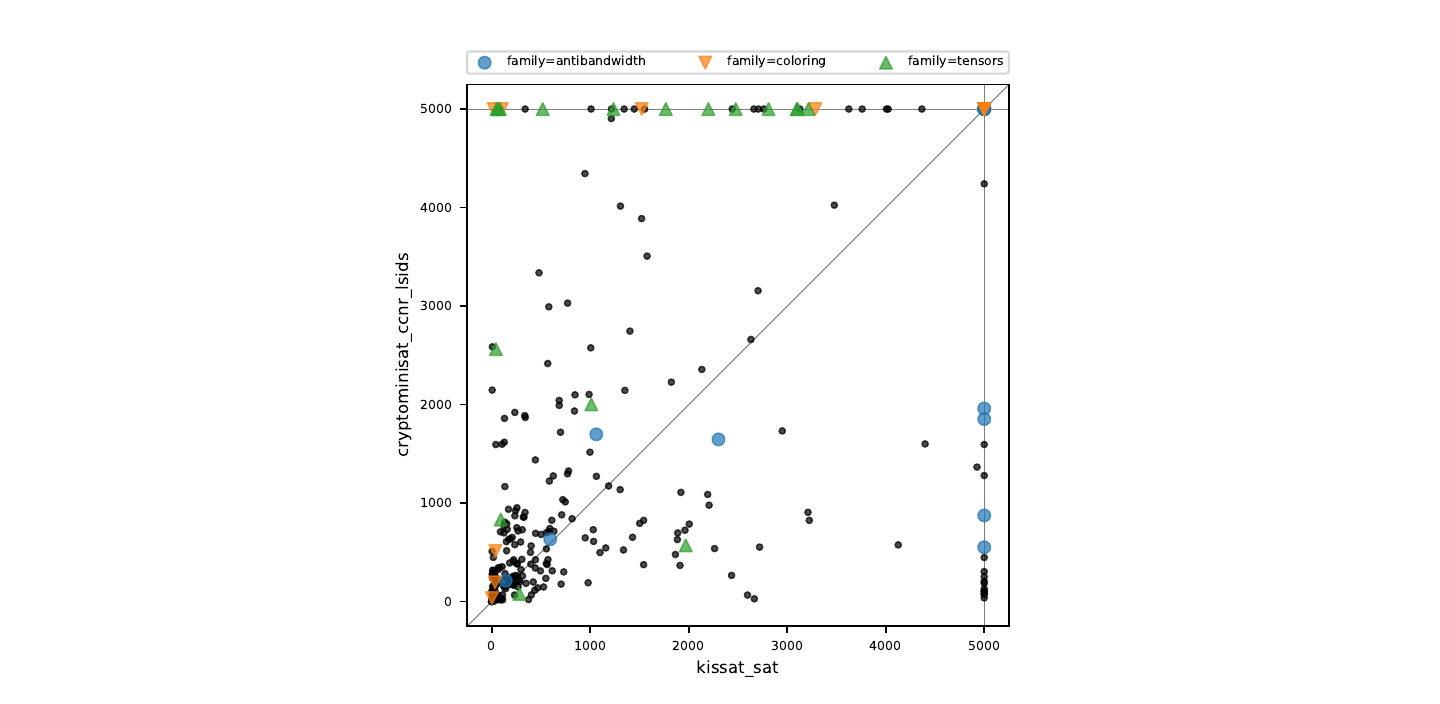}
\caption{Scatter plot of Kissat and Cryptominisat runtimes in SAT Competition~2020 highlighting the three specified families antibandwidth, coloring and tensors, and leaving the other families black}
\label{fig:plot}
\end{figure}
For producing \emph{augmented} runtime plots, \textsf{gbd} offers the \textsf{plot} command which allows us to use \textsf{GBD Queries} to highlight instances of specific properties. 
The following example shows the creation of an augmented scatter plot with the runtimes of Kissat and Cryptominisat in SAT~Competition~2020. 
In order to highlight the complementary strengths of those two solvers, we specify a list of sub-queries to filter for specific instance families. 
The resulting plot is presented in Figure~\ref{fig:plot}.

\begin{shell}
%prompt gbd plot scatter "competition_track = main_2020" 
	-r kissat_sat cryptominisat_ccnr_lsids 
	-g family=antibandwidth family=coloring family=tensors
\end{shell}
%
%
%\newpage
\section{GBD Web Interface}
\label{sec:web}

\textsf{GBD Tools} also comes with the web-interface \textsf{gbd-server} that supports access to instances and data via \textsf{GBD Queries}. 
Our public \textsf{gbd-server} distributes benchmark instances and feature databases of SAT Competitions.\footnote{\url{https://gbd.iti.kit.edu}} 
In addition to its web-interface, \texttt{gbd-server} exposes a couple of micro-services (cf. Table~\ref{tab:services}). 
For example, we can use these micro-services to download instances as follows. 

\begin{shell}
%prompt wget --content-disposition http://gbd.iti.kit.edu/file/<gbdhash>
\end{shell}
%

%\begin{figure}[t]
%\centering
%\includegraphics[width=\linewidth, trim=0 30 0 36, clip]{screenshot.png}
%\caption[GBD Web Interface (Screenshot)]{Screenshot of GBD Website \url{https://gbd.iti.kit.edu}}
%\label{fig:gbd-web}
%\end{figure}

\begin{table}[t]
\centering
\begin{tabular}{ll}
{\bf Attribute Value} ~&~ \texttt{http://gbd.iti.kit.edu/attribute/$\langle$name$\rangle$/$\langle$gbd-hash$\rangle$}\\
{\bf All Attribute Values} ~&~ \texttt{http://gbd.iti.kit.edu/info/$\langle$gbd-hash$\rangle$}\\
{\bf Download Instance} ~&~ \texttt{http://gbd.iti.kit.edu/file/$\langle$gbd-hash$\rangle$}
\end{tabular}~\\[.7em]
\caption{URI Schemes of GBD Micro-Services}
\label{tab:services}
\end{table}

\section{Conclusion}
\label{sec:conclusion}

Our approach benefits largely from our solution to the \emph{instance identification} problem which has not satisfactorily been solved in previous approaches. 
\textsf{GBD Hash} facilitates instance distribution with a data-driven catalog, and experiment evaluation with augmented runtime data. 
\textsf{GBD Tools} are recently used for compiling and distributing benchmark instances for SAT competitions~\cite{SC2020:AIJ}. 
We are extending \textsf{GBD Tools} with methods for differentiated analysis of small portfolios and runtime prediction models. 
That includes efficient feature extraction for bootstrapping of data and extended support for other data-formats such as Aslib. 
Our tool can be extended to mangage other types of instances as well by defining a dedicated instance identifier, e.g., for QBF, MaxSAT, or CSP.

\bibliography{main}

\end{document}